\documentclass[graybox]{svmult}

\usepackage{type1cm}        
\usepackage{makeidx}         
\usepackage{graphicx}        
\usepackage{multicol}        
\usepackage[bottom]{footmisc}

\usepackage{newtxtext}       %
\usepackage[varvw]{newtxmath}       

\usepackage[mathscr]{eucal}
\usepackage{mathtools}
\DeclarePairedDelimiter\abs{\lvert}{\rvert}%

\makeindex             

\begin{document}

\title*{End-to-End Image-Based Fashion Recommendation}
\author{Shereen Elsayed, Lukas Brinkmeyer and Lars~Schmidt-Thieme}
\institute{Shereen Elsayed \at Information Systems and Machine Learning Lab, University of Hildesheim, Germany \email{elsayed@ismll.uni-hildesheim.de}
\and Lukas Brinkmeyer \at Information Systems and Machine Learning Lab, University of Hildesheim, Germany \email{brinkmeyer@ismll.uni-hildesheim.de}
\and Lars~Schmidt-Thieme \at Information Systems and Machine Learning Lab, University of Hildesheim, Germany \email{schmidt-thieme@ismll.uni-hildesheim.de}}

%
\maketitle

\abstract{In fashion-based recommendation settings, incorporating the item image features is considered a crucial factor, and it has shown significant improvements to many traditional models, including but not limited to matrix factorization, auto-encoders, and nearest neighbor models. While there are numerous image-based recommender approaches that utilize dedicated deep neural networks, comparisons to attribute-aware models are often disregarded despite their ability to be easily extended to leverage items' image features. In this paper, we propose a simple yet effective attribute-aware model that incorporates image features for better item representation learning in item recommendation tasks. The proposed model utilizes items' image features extracted by a calibrated ResNet50 component. We present an ablation study to compare incorporating the image features using three different techniques into the recommender system component that can seamlessly leverage any available items' attributes. Experiments on two image-based real-world recommender systems datasets show that the proposed model significantly outperforms all state-of-the-art image-based models.}

\section{Introduction}
\label{sec:1}
In recent years, recommender systems became one of the essential areas in the machine learning field. In our daily life, recommender systems affect us in one way or another, as they exist in almost all the current websites such as social media, shopping websites, and entertainment websites. Buying clothes online became a widespread trend nowadays; websites such as Zalando and Amazon are getting bigger every day. In this regard, item images play a crucial role, no one wants to buy a new shirt without seeing how it looks like. Many users also can have the same taste when they select what they want to buy; for example, one loves dark colors most of the time, others like sportswear, and so on.
In fashion recommendation adding the items images to the model has proven to show a significant lift in the recommendation performance when the model is trained not only on the relational data describing the interaction between users and items but also on how it looks. It can massively help the model recommend other items that look similar or compatible. Some recent works tackled this area in the last years, particularly adding item images into the model with different techniques.

Following this research direction, we propose a hybrid attribute-aware model that relies on adding the items' image features into the recommendation model. 

The contributions through this works can be specified as follows;
\begin{itemize}
\item We propose a simple image-aware model for item recommendation that can leverage items' image features extracted by a fine-tuned ResNet50 component \cite{he2016deep}.
\item We conducted extensive experiments on two benchmark datasets, and results show that the proposed model was able to outperform more complex state-of-the-art methods.
\item We conduct an ablation study to analyze and compare three different approaches to include the items images features extracted using the ReNet50 into the recommender model.
\end{itemize}

\section{Related work}
\label{sec:2}

Many image-based recommender systems were proposed in the last few years. They are becoming more popular, and their applications are getting wider, especially in fashion e-commerce websites. Many of the proposed models in the literature relied on using \textbf{pre-trained networks} for items images features extraction. In 2016 an essential state-of-the-art model was proposed, VBPR \cite{he2015vbpr}. It uses the BPR ranking model \cite{rendle2012bpr} for prediction and takes the visual item features into account. They use a pre-trained CNN network to extract the item features; these features pass through a fully connected layer to obtain the latent embedding. 
Another model which uses more than one type of external information is JRL \cite{zhang2017joint}. It incorporates three different information sources (user reviews, item images, and ratings) using a pre-trained CaffeNet and PV-DBOW model \cite{le2014distributed}.  
While in 2017, Qiang et al. \cite{liu2017deepstyle} proposed the DeepStyle model, where their primary assumption is that the item representation consists of style and item category representations extracting the item image features via a CaffeNet model. To get the style features, they subtract latent factors representing the category. Additionally, a region-guided approach (SAERS) \cite{hou2019explainable} introduced the items' visual features using AlexNet to get general features and utilizing a ResNet-50 architecture for extracting semantic features representing the region of interest in the item image. Before semantic features are added to the global features, an attention mechanism using the users' preferences is applied. The final item embedding becomes the semantic features combined with the global features. 

The image networks used for item image feature extraction can also be \textbf{trained end-to-end} with the recommender model; the most popular model applied this technique is the DVBPR \cite{kang2017visually} powerful model proposed in 2017 that incorporates visual item features. It does two tasks; the first task is training the BPR recommender model jointly with a CNN structure to extract the item's image pixel-level features. The second task uses Generative Adversarial Networks (GANs) to generate new item images based on user preferences.

Attribute-aware recommender system models are a family of hybrid models that can incorporate external user and item attributes. Theoretically, some of these models can be extendable to image-based settings by carefully converting the raw image features into real-valued latent features, which can be used as the item's attributes. Recently,  (CoNCARS) \cite{costa2019collective} model was proposed that takes the user and item one-hot- encoded vectors as well as the user/item timestamps. The mode utilizes a convolution neural network (CNN) on top of the interaction matrix to generate the latent embeddings. Parallel work by Rashed et al. proposed an attribute-aware model (GraphRec) \cite{rashed2019attribute} that appends all the users' and items' attributes on-hot-encoded vectors. It extracts the embeddings through neural network layers that can capture the non-linearity in the user-item relation.

In the literature, using attribute-aware models has been mainly set aside for image-based items ranking problems. Hence, in this paper, we propose a simple image-aware model that utilizes the latent image features as item attributes. The experimental results show that the proposed model outperforms current complex image-based state-of-the-art models.

%
%
%
\section{Methodology}
\label{sec:3}
\subsection{Problem Definition}
In image-based item recommendation tasks, there exist a set of $M$ users $\mathcal{U}:=\lbrace{u_1, \cdots, u_M}\rbrace$, a set of $N$ items $\mathcal{I}:=\lbrace{i_1, \cdots, i_N}\rbrace$ with their images $X_i \in \mathbb{R}^{N \times (L \times H \times C)}$ of dimensions $L \times H$ and $C$ channels,  and a sparse binary interaction matrix $R \in \mathbb{R}^{M \times N}$ that indicate user's implicit preferences on items based on historical interactions.

 The recommendation task's primary goal is to generate a ranked personalized short-list of items to users by estimating the missing likelihood scores in $R$ while considering the visual information that exists in the item images.

\subsection{Proposed model}
The proposed model consists of an image features extraction component and an attribute-aware recommender system component that are jointly optimized. 

\subsubsection{Recommender System Component}
Inspired by the GraphRec model \cite{rashed2019attribute}, the recommender system component utilizes the user's one-hot encoded input vector and concatenates the external features of the items directly to the items' one-hot input vectors. These vectors are then fed to their independent embedding functions $\psi_u$ : $\mathbb{R}^{M}  \rightarrow \mathbb{R}^{K}$ and $\psi_i$ : $\mathbb{R}^{(N + F)}  \rightarrow \mathbb{R}^{K}$ as follows:

\begin{equation}
    z_{u}= \psi_u(v_u)=  v_u W^{\psi_u} + b^{\psi_u}
\end{equation}

\begin{equation}
    z_{i}= \psi_i(v_i)= concat(v_i, \phi(x_i)) W^{\psi_i} + b^{\psi_i}
\end{equation}

\noindent where $W^{\psi_u}$, $W^{\psi_i}$ are the weight matrices of the embedding functions, and $b^{\psi_u}$, $b^{\psi_i}$ are the bias vectors. $v_u$, $v_i$ represents the user and item one-hot encoded vectors. Additionally,  $\phi(x_i)$ represents the features extraction component that embeds an item's raw image $x_i$ to a latent feature vector of size $F$.

After obtaining the user and item embeddings, the final score is calculated using the dot-product of the two embedding vectors, $\hat{y}_{u i}= z_u \cdot z_i$ to give the final output score representing how much this user $u$ will tend to like this item $i$. The final score is computed via a sigmoid function; $\sigma(\hat{y_{ui}})=1/1+e^{(\hat{y_{ui}})}$, to limit the score value from $0\rightarrow 1$. The model target is defined as $y_{u i}$ which is for implicit feedback either $0$ or $1$;
\begin{equation}
    y_{u i}=
    \begin{cases}
      1, & \text{observed item}; \\
      0, & \text{otherwise}                      
    \end{cases}
\end{equation}
Given the users-items positive and negative interactions $D_s^+$, $D_s^-$, and the output score $\hat{y}_{u i}$ of the model and the original target $y_{ui}$, we use negative sampling for generating unobserved instances and optimize the negative log-likelihood objective function $\ell(\hat{y}; D_s)$ using ADAM optimizer, which can be defined as;
\begin{equation}
 \label{eq:5}
       - \sum_{(u,i)\in {D_s^+ \bigcup D_s^- }} y_{u i} \log{(\hat{y}_{u i})} + (1-y_{u i}) (1-\log{(\hat{y}_{u i})})
\end{equation}

\subsubsection{Extraction of Image Features}
\begin{figure}[ht]
  \centering
  \includegraphics[scale=0.8]{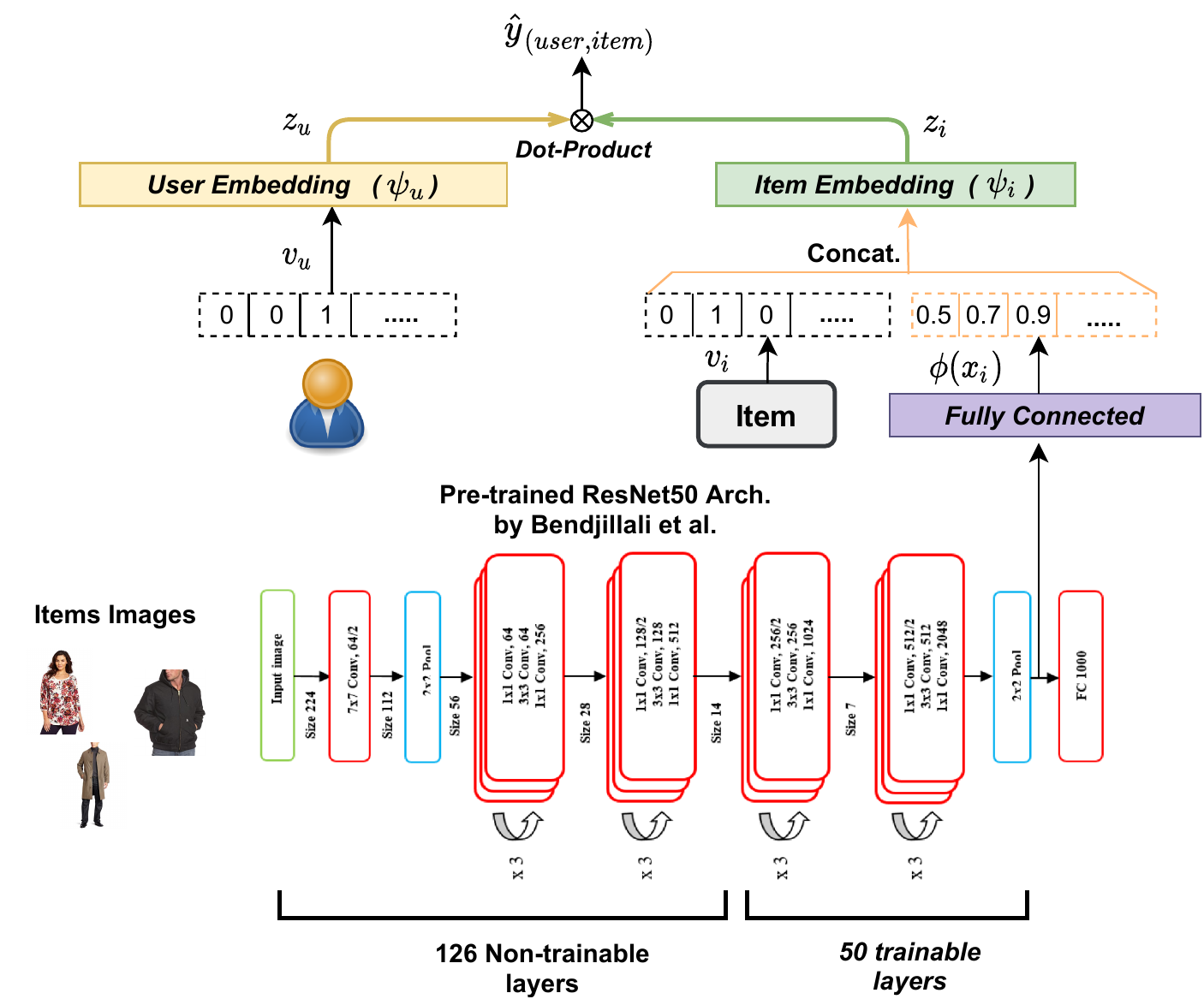}
  \caption{End-to-End ImgRec Architecture}
  \label{fig:3}
\end{figure}

To extract the latent item's image features, we propose using the ResNet50 component for combining the raw image features. To refine the image features further and get better representation, we can jointly train the whole image network simultaneously with the recommender model. However, it will require a considerable amount of memory and computational power to load and update these parameters. In this case, ResNet50 consists of 176 layers with around 23 million parameters. To mitigate this problem, we propose ImgRec End-to-End (ImgRec-EtE), where we utilize a ResNet50 \cite{he2016deep} pre-trained on ImageNet dataset \cite{imagenet_cvpr09} and jointly train part of the image network to be updated with the recommender model, and at the same time, benefit from starting with the pre-trained weights. As shown in Figure \ref{fig:3}, we selected the last 50 layers to be updated and fix the first 126 layers.
Furthermore, we added an additional separate, fully connected layer to fine-tune the image features extracted by the ResNet50. This layer will be trained simultaneously with the recommender model. Moreover, this additional layer makes the image features more compact and decreases its dimensionality further to match the user latent embedding. Thus the features extraction function $\phi(x_i)$ for ImgRec-EtE can be defined as follows; 
\begin{equation}
    \phi(x_i) := ReLU(ResNet50(x_i) W^{\phi} + b^{\phi})
\end{equation}

\subsection{Training Strategy}
To increase the speed of the training process, we used a two-stage training protocol. Firstly, we train the model by fixing the image network pre-trained parameters and updating the recommender model parameters until convergence. After obtaining the best model performance in the first stage, we jointly learn and fine-tune the last 50 layers of the image network further with the recommender model. This methodology allowed us to fine-tune the model after reaching the best performance given the pre-trained image network weights, also it saves time and computational power while achieve superior prediction performance compared to using only the fixed pre-trained parameters of the image network.

\section{Experiments}
Through the experimental section, we aim to answer the following research questions:\\
\textbf{RQ1} How does the proposed model fair against state-of-the-art image-based models?\\
\textbf{RQ2} What is the best method for adding images features to the model? 
\subsection{Datasets}

We chose two widely used image-based recommendation datasets \textit{Amazon fashion} and \textit{Amazon men}, introduced by McAuley et al. \cite{mcauley2015image}. \textit{Amazon fashion} was collected from six different categories, "men/women's tops, bottoms, and shoes," while \textit{Amazon men} contains all subcategories (gloves, scarves, sunglasses, etc.). The number we mention of users-items of the Fashion dataset is different from the ones stated in the main paper. However, we contacted the authors\footnote{https://github.com/kang205/DVBPR/issues/6} and the numbers in Table \ref{tab:2} were found to be the correct ones.
%

\begin{table}[!t]
\caption{Datasets statistics}
\label{tab:2}       
%
%
\begin{tabular}{p{2.3cm}p{1.4cm}p{1.4cm}p{1.4cm}p{1.4cm}}
\hline\noalign{\smallskip}
Dataset& Users & Items & Categories& Interactions\\
\noalign{\smallskip}\svhline\noalign{\smallskip}
Amazon fashion& 45184 & 166270 & 6 & 267635  \\
Amazon men& 34244 & 110636 & 50 & 186382 \\
\noalign{\smallskip}\hline\noalign{\smallskip}
\end{tabular}
\end{table}

\subsection{Evaluation Protocol}
To evaluate our proposed models, the data is split into training, validation, and test sets using the leave-one-out evaluation protocol as in \cite{kang2017visually, hou2019explainable,he2015vbpr}. However, we used two different numbers of negative samples in the evaluation protocol for the direct comparison against the published results of the state-of-the-art models DVBPR \cite{kang2017visually}, and SAERS \cite{hou2019explainable} because both models originally used a different number of samples, and the source code of SAERS is not available. 

For the direct comparison against DVBPR, we sample 100 negative samples ($\mathcal{I}^t$) and one positive item $i$ for each user. On the other hand, for direct comparison against our second baseline SAERS \cite{hou2019explainable} we sample 500 negative items ($\mathcal{I}^t$) and one positive item $i$. To ensure our results' consistency, we report the mean of each experiment's five different trials. 
 
For evaluation, we report the \textbf{Area Under the Curve (AUC)} as it is the primary metric in all of our baselines' papers, further more it is persistent to the number of negative samples used \cite{krichene2020sampled}: 
\begin{equation}
 \label{eq:9}
     AUC= \frac{1}{\abs{\mathcal{U}}} \sum_{u\in \mathcal{U} } \frac{1}{\abs{\mathcal{I}^t}} \sum_{i, j\in \mathcal{I}^t} (\hat{y}_{u i} > \hat{y}_{u j})
\end{equation}
\begin{table}[!t]
\caption{Comparison of AUC scores with 100 negative samples per user, the bold results represent the best performing model and we underline the second best result.}
\label{tab:3}       
%
%
\begin{tabular}{p{2.3cm}p{1.45cm}p{1.4cm}p{1.45cm}p{1.4cm}}
\hline\noalign{\smallskip}
Datasets &\multicolumn{4}{c}{\textit{Interactions}}\\
         &PopRank&WARP&BPR-MF&FM\\
\noalign{\smallskip}\svhline\noalign{\smallskip}
\textit{Amazon Fashion}&0.5849&0.6065&0.6278&0.7093\\
\textit{Amazon Men} &0.6060&0.6081&0.6450&0.6654\\
\noalign{\smallskip}\hline\noalign{\smallskip}
\end{tabular}

\begin{tabular}{p{2.3cm}p{1.4cm}p{1.4cm}p{1.4cm}p{1.5cm}}
\hline\noalign{\smallskip}
Datasets &\multicolumn{4}{c}{\textit{Interactions+Image Features}}\\
         &VisRank&VBPR&DVBPR& ImgRec-EtE\\
\noalign{\smallskip}\svhline\noalign{\smallskip}
\textit{Amazon Fashion}&0.6839&0.7479&\underline{0.7964}&\textbf{0.8250}\\
\textit{Amazon Men} &0.6589&0.7089&\underline{0.7410}&\textbf{0.7899}\\
\noalign{\smallskip}\hline\noalign{\smallskip}
\end{tabular}
\end{table}
\begin{table}[!t]
\caption{Comparison of AUC scores with 500 negative samples per user.}
\label{tab:4}       
%
%
\begin{tabular}{p{2.3cm}p{1.2cm}p{1.2cm}|p{1.1cm}p{1.3cm}p{1.1cm}p{1.2cm}p{1.5cm}}
\hline\noalign{\smallskip}
Datasets &\multicolumn{2}{c}{\textit{Interactions}}&\multicolumn{5}{c}{\textit{Interactions+Image Features}}\\
         &PopRank&BPR-MF&VBPR&DeepStyle&JRL&SAERS&ImgRec-EtE\\
\noalign{\smallskip}\svhline\noalign{\smallskip}
\textit{Amazon Fashion}&0.5910&0.6300&0.7710&0.7600&0.7710&\underline{0.8161}&\textbf{0.8250}\\
\noalign{\smallskip}\hline\noalign{\smallskip}
\end{tabular}
\end{table}

\subsection{Baselines}
We compared our proposed methods to the published results of the state-of-the-art image-based models DVBPR and SAERS. We also compared our results against a set of well-known item recommendation models that were used in \cite{kang2017visually,hou2019explainable}.
\begin{itemize}
\item \textbf{PopRank}: A naive popularity-based ranking model.
\item \textbf{WARP \cite{weston2011wsabie}}: A matrix factorization model that uses Weighted Approximate-Rank Pairwise (WARP) loss.
\item \textbf{BPR-MF \cite{rendle2012bpr}}: A matrix factorization model that uses the BPR loss to get the ranking of the items.
\item \textbf{VisRank}: A content-based model that utilizes the similarity between CNN features of the items bought by the user.

\item \textbf{Factorization Machines (FM) \cite{rendle2010factorization}}: A generic method that combines the benefits of both SVM and factorization techniques using pair-wise BPR loss.
\item \textbf{VBPR \cite{he2015vbpr}}: A model that utilizes items; visual features, using pre-trained CaffeNet and a BPR ranking model.
\item \textbf{DeepStyle \cite{liu2017deepstyle}}: A model that uses the BPR framework and incorporates style features extracted by subtracting category information from CaffeNet visual features of the items.
\item \textbf{JRL\cite{zhang2017joint}}: A model that incorporates three different types of item attributes. In this case, we considered only the visual features for comparison purposes.
\item \textbf{DVBPR \cite{kang2017visually}}: State-of-the-art image-based model that adds the visual features extracted from a dedicated CNN network trained along with a BPR recommender model.
\item \textbf{SAERS \cite{hou2019explainable}}: State-of-the-art image-based model that utilizes the region of interests in the visual images of the items while also considering the global extracted features from a dedicated CNN to get the final items representations. 
\end{itemize}
\subsection{Comparative study against state-of-the-art image-based models (RQ1)}
Since the DVBPR baseline conducted their results on Amazon men and Amazon fashion datasets, we compared our results directly to both datasets' published results. On the other hand, the SAERS model used only the Amazon fashion dataset, so we only report the results for this dataset using 500 negative samples per user. Table \ref{tab:3} illustrates the ImgRec-EtE obtained results against VBPR and DVBPR results. The proposed model ImgRec-EtE represents the best performance on both men and fashion datasets. It shows a 2.5\% improvement over the fashion dataset DVBPR reported performance and a 4.8\% improvement for the men dataset. The results show consistent AUC values regardless of the number of negative samples, as per the recent study by Krichene et al. \cite{krichene2020sampled}.
Table \ref{tab:4} demonstrates the comparison against the DeepStyle, JRL and SAERS models. The proposed model ImgRec-EtE represents the best performance on the fashion dataset. Despite its simplicity, the model has achieved an AUC of 0.825, which shows an improvement over the complex state-or-the-art SAERS model with 0.9\%. 
\subsection{Ablation Study (RQ2) }
Besides obtaining the items features in an end-to-end fashion, it is worth mentioning that we tried other methods to incorporate the images' features. Firstly in (ImgRec-Dir), we directly concatenate the image features extracted using the output of the next to last fully connected layer of a pre-trained ResNet50 to the one-hot encoded vector representing the item. 
On the other hand (ImgRec-FT) passes the features extracted using the pre-trained network to a fine-tuning layer that is trained with the recommender model and obtain better item representation. Subsequently, the item's image latent features are concatenated to the item one-hot encoded vector to form one input vector representing the item. As shown in Table \ref{tab:5} The images' features had a varying effect depending on how they were added to the model; ImgRec-Dir achieved an AUC of 0.77 on the Amazon fashion dataset and 0.736 on the Amazon men dataset. While looking into ImgRec-FT performance after adding the fine-tuning layer, we can see an improvement of 3.2\% on the Amazon fashion dataset and 1.6\% on the Amazon men dataset performances, which shows high competitiveness against the state-or-the-art models while having a much lower computational complexity. Finally, ImgRec-EtE, which jointly trains part of the ResNet50 simultaneously with the model, positively impacted the results with further improvement of 1.6\%  over the ImgRec-FT performance on both datasets. 

\begin{table}[!t]
\caption{Comparison of AUC scores with 100 negative samples per user, between the three ways of incorporating the image features.}
\label{tab:5}       
%
%
\begin{tabular}{p{2.3cm}p{1.6cm}p{1.6cm}p{1.6cm}}
\hline\noalign{\smallskip}
Datasets & ImgRec-Dir & ImgRec-FT & ImgRec-EtE\\
\noalign{\smallskip}\svhline\noalign{\smallskip}
\textit{Amazon Fashion}&0.7770&0.8090&\textbf{0.8250}\\
    \textit{Amazon Men} &0.7363&0.7735&\textbf{0.7899}\\
\noalign{\smallskip}\hline\noalign{\smallskip}
\end{tabular}
\end{table}

\subsection{Hyperparameters}
We ran our experiments using GPU RTX 2070 Super and CPU Xeon Gold 6230 with RAM 256 GB. We used user and item embedding sizes of 10 and 20 with \textit{Linear} activation function for both datasets. We applied grid search on the learning rate between [0.00005 and 0.0003] and the L2-regularization lambda between [0.000001 and 0.2]. The best parameters are 0.0001 and 0.1 for ImgRec-Dir and ImgRec-FT. While in ImgRec-EtE case, the best L2-regularization lambda is 0.000001 for phase1 (fixed-weights) and 0.00005 for phase 2 (joint training). The features fine-tuning layer, the best-selected embedding size is 150 with \textit{ReLU} activation function. ImgRec codes and datasets are available at https://github.com/Shereen-Elsayed/ImgRec.

\section{Conclusion}
In this work, we propose an image-based attribute-aware model for items' personalized ranking with jointly training a ResNet50 component simultaneously with the model for incorporating image features into the recommender model. Adding the image features showed significant improvement in the model's performance. ImgRec-EtE shows superior performance to all image-based recommendation approaches. Furthermore, we conducted an ablation study to compare different approaches of adding the features to the model; direct features concatenation, adding a fine-tuning fully connected layer, and jointly training part of the image network. 

\section{Acknowledgements}
This work is co-funded by the industry Project “IIP-Ecosphere: Next Level Ecosphere for Intelligent Industrial Production”.
\bibliographystyle{acm}
\bibliography{ImgRec}

\begin{thebibliography}{10}

\bibitem{costa2019collective}
{\sc Costa, F. S.~d., and Dolog, P.}
\newblock Collective embedding for neural context-aware recommender systems.
\newblock In {\em Proceedings of the 13th ACM Conference on Recommender
  Systems\/} (2019), pp.~201--209.

\bibitem{imagenet_cvpr09}
{\sc Deng, J., Dong, W., Socher, R., Li, L.-J., Li, K., and Fei-Fei, L.}
\newblock {ImageNet: A Large-Scale Hierarchical Image Database}.
\newblock In {\em CVPR09\/} (2009).

\bibitem{he2016deep}
{\sc He, K., Zhang, X., Ren, S., and Sun, J.}
\newblock Deep residual learning for image recognition.
\newblock In {\em Proceedings of the IEEE conference on computer vision and
  pattern recognition\/} (2016), pp.~770--778.

\bibitem{he2015vbpr}
{\sc He, R., and McAuley, J.}
\newblock Vbpr: visual bayesian personalized ranking from implicit feedback.
\newblock {\em arXiv preprint arXiv:1510.01784\/} (2015).

\bibitem{hou2019explainable}
{\sc Hou, M., Wu, L., Chen, E., Li, Z., Zheng, V.~W., and Liu, Q.}
\newblock Explainable fashion recommendation: A semantic attribute region
  guided approach.
\newblock {\em arXiv preprint arXiv:1905.12862\/} (2019).

\bibitem{kang2017visually}
{\sc Kang, W.-C., Fang, C., Wang, Z., and McAuley, J.}
\newblock Visually-aware fashion recommendation and design with generative
  image models.
\newblock In {\em 2017 IEEE International Conference on Data Mining (ICDM)\/}
  (2017), IEEE, pp.~207--216.

\bibitem{krichene2020sampled}
{\sc Krichene, W., and Rendle, S.}
\newblock On sampled metrics for item recommendation.
\newblock In {\em Proceedings of the 26th ACM SIGKDD International Conference
  on Knowledge Discovery \& Data Mining\/} (2020), pp.~1748--1757.

\bibitem{le2014distributed}
{\sc Le, Q., and Mikolov, T.}
\newblock Distributed representations of sentences and documents.
\newblock In {\em International conference on machine learning\/} (2014),
  pp.~1188--1196.

\bibitem{liu2017deepstyle}
{\sc Liu, Q., Wu, S., and Wang, L.}
\newblock Deepstyle: Learning user preferences for visual recommendation.
\newblock In {\em Proceedings of the 40th International ACM SIGIR Conference on
  Research and Development in Information Retrieval\/} (2017), pp.~841--844.

\bibitem{mcauley2015image}
{\sc McAuley, J., Targett, C., Shi, Q., and Van Den~Hengel, A.}
\newblock Image-based recommendations on styles and substitutes.
\newblock In {\em Proceedings of the 38th international ACM SIGIR conference on
  research and development in information retrieval\/} (2015), pp.~43--52.

\bibitem{rashed2019attribute}
{\sc Rashed, A., Grabocka, J., and Schmidt-Thieme, L.}
\newblock Attribute-aware non-linear co-embeddings of graph features.
\newblock In {\em Proceedings of the 13th ACM Conference on Recommender
  Systems\/} (2019), pp.~314--321.

\bibitem{rendle2010factorization}
{\sc Rendle, S.}
\newblock Factorization machines, icdm, 2010.

\bibitem{rendle2012bpr}
{\sc Rendle, S., Freudenthaler, C., Gantner, Z., and Schmidt-Thieme, L.}
\newblock Bpr: Bayesian personalized ranking from implicit feedback.
\newblock {\em arXiv preprint arXiv:1205.2618\/} (2012).

\bibitem{weston2011wsabie}
{\sc Weston, J., Bengio, S., and Usunier, N.}
\newblock Wsabie: Scaling up to large vocabulary image annotation.

\bibitem{zhang2017joint}
{\sc Zhang, Y., Ai, Q., Chen, X., and Croft, W.~B.}
\newblock Joint representation learning for top-n recommendation with
  heterogeneous information sources.
\newblock In {\em Proceedings of the 2017 ACM on Conference on Information and
  Knowledge Management\/} (2017), pp.~1449--1458.

\end{thebibliography}
\end{document}